\documentclass[10pt,twocolumn,letterpaper]{article}

\usepackage{cvpr}
\usepackage{times}
\usepackage{epsfig}
\usepackage{graphicx}
\usepackage{amsmath}
\usepackage{amssymb}

\cvprfinalcopy 
\def\cvprPaperID{24} 

\ifcvprfinal\fi

\title{Exploring Factors for Improving Low Resolution Face Recognition}
\author{Omid Abdollahi Aghdam\textsuperscript{1}
\thinspace Behzad Bozorgtabar\textsuperscript{2} \\ 
Haz{\i}m Kemal Ekenel\textsuperscript{1} 
\thinspace Jean-Philippe Thiran\textsuperscript{2} \\ 
\textsuperscript{1} SiMiT Lab, ITU, Turkey 
\thinspace \textsuperscript{2} LTS5, EPFL, Switzerland \\
\tt\small abdollahi15@itu.edu.tr \thinspace behzad.bozorgtabar@epfl.ch \\ \tt\small ekenel@itu.edu.tr \thinspace jean-philippe.thiran@epfl.ch}

\begin{document}
\maketitle
\begin{abstract}
   State-of-the-art deep face recognition approaches report near perfect performance on popular benchmarks, e.g., Labeled Faces in the Wild. However, their performance deteriorates significantly when they are applied on low quality images, such as those acquired by surveillance cameras. A further challenge for low resolution face recognition for surveillance applications is the matching of recorded low resolution probe face images with high resolution reference images, which could be the case in watchlist scenarios. In this paper, we have addressed these problems and investigated the factors that would contribute to the identification performance of the state-of-the-art deep face recognition models when they are applied to low resolution face recognition under mismatched conditions. We have observed that the following factors affect performance in a positive way: appearance variety and resolution distribution of the training dataset, resolution matching between the gallery and probe images, and the amount of information included in the probe images. By leveraging this information, we have utilized deep face models trained on MS-Celeb-1M and fine-tuned on VGGFace2 dataset and achieved state-of-the-art accuracies on the SCFace and ICB-RW benchmarks, even without using any training data from the datasets of these benchmarks.
\end{abstract}

\section{Introduction}

\begin{figure}
 \begin{center}
 \includegraphics[width=\linewidth,keepaspectratio=true]{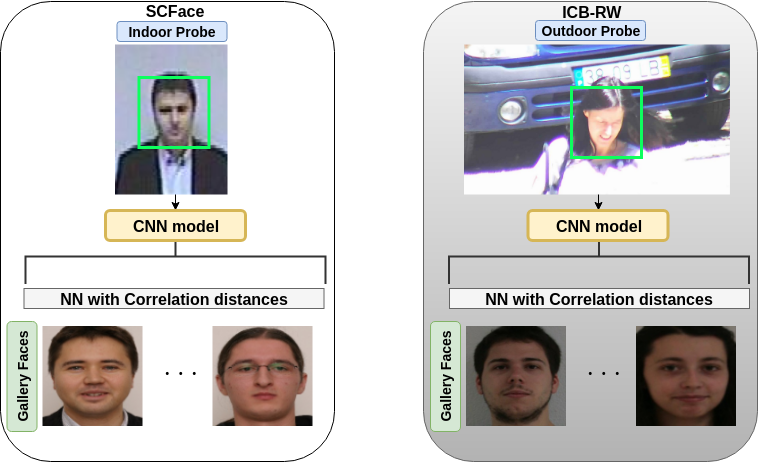}
 \end{center}
 \caption{Face identification scenario addressed in this paper. The scenario resembles a watchlist one, in which we have high quality gallery face images of the individuals recorded at indoor studio settings and low quality probe face images recorded by indoor, as in the SCFace benchmark~\cite{grgic2011scface} (left), or outdoor, as in the ICB-RW benchmark~\cite{neves2016icb} (right), surveillance cameras.}
 \label{fig:protocol}
\end{figure}

Face recognition systems are now very common, from applications in our smartphones to security gates in the airports. These systems work flawlessly, when the training and test images are of high quality, have similar distributions, and do not vary much. However, in the surveillance scenarios, in which the training and test images do not have the same distribution, face recognition systems' performance deteriorates. Figure \ref{fig:protocol} illustrates the face identification scenario addressed in this paper to explore this problem. The scenario resembles a watchlist one, in which we have high quality gallery face images of the individuals recorded at indoor studio settings and low quality probe face images recorded by indoor, as in the SCFace~\cite{grgic2011scface}, or outdoor, as in the ICB-RW~\cite{neves2016icb}, surveillance cameras. 

The recent breakthroughs in deep learning architectures~\cite{hu2018squeeze, he2016deep, szegedy2015going, simonyan15} and availability of large-scale training databases, e.g. CASIA Webface~\cite{Yi2014LearningFR}, MS-Celeb-1M~\cite{guo2016ms}, VGGFace2~\cite{cao2018vggface2}, have aided the research in face recognition (FR). The advancements have been significant on the benchmarks that have relatively high resolution face images in gallery and probe sets, e.g. Labeled Faces in the Wild (LFW)~\cite{huang2008labeled} and YouTube Faces (YTF)~\cite{wolf2011face}.

\begin{figure}[hbt!]
 \begin{center}
 \includegraphics[width=\linewidth,keepaspectratio=true]{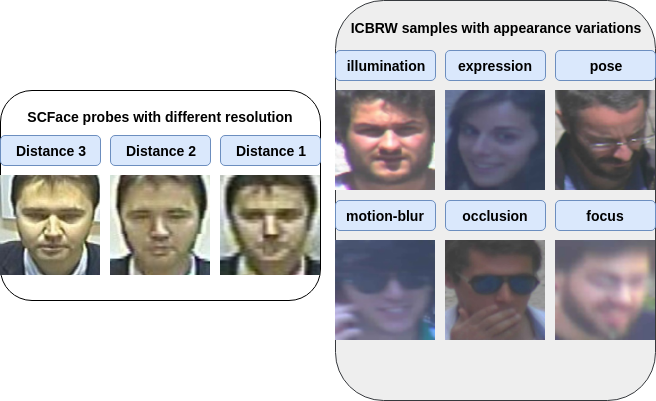}
 \end{center}
 \caption{Sample probe images from the SCFace and the ICB-RW datasets. The probe set of SCFace contains face images with three different resolutions depending on the distance between the subject and the cameras (left), probe set of ICB-RW includes face images recorded outdoors and contain  challenging appearance variations (right).}
 \label{fig:probe-samples}
\end{figure}

In low resolution face recognition under surveillance scenarios on the other hand, for example in a watchlist application, there is a single high resolution frontal face image per subject in the gallery set, whereas, there are low resolution face images captured with surveillance cameras in the probe set, which contain appearance variations due to changes in illumination, expression, pose, motion-blur, occlusion, focus, and varying resolutions as can be seen in Figure \ref{fig:probe-samples} for the ICB-RW \cite{neves2016icb} and the SCFace \cite{grgic2011scface} benchmarks. Probe face images' quality problems and the quality mismatch between the gallery and probe images are the main causes of the performance drop in deep face recognition models when they are tested under such conditions.\par

To address the challenges posed by low resolution face recognition, in this work, we explore the factors that would improve Low Resolution Face Recognition (LRFR) performance. We investigate the factors, such as, appearance variety and resolution distribution of the training database, resolution matching between the gallery and probe images, and the amount of information included in the probe images. We observe that all of these factors improve the performance. We test the robustness of four state-of-the-art deep convolutional neural network (CNN) models, namely, ResNet-50~\cite{he2016deep}, SENet-50~\cite{hu2018squeeze}, LResNet50E-IR~\cite{deng2018arcface}, LResNet100E-IR~\cite{deng2018arcface} and utilize two large scale face databases, VGGFace2~\cite{cao2018vggface2} and MS-Celeb-1M~\cite{guo2016ms}, to train and fine-tune them. We present that appearance variety and resolution distribution of the training database is of paramount importance. We also analyze the impact of the resolution matching between the gallery and probe images. In contrast to a previous work~\cite{wang2016studying}, instead of super-resolving low resolution face images to match the resolution of the gallery and probe images, we down-sample the high resolution gallery images. We observe that matching the gallery and probe face images' resolution increases the performance significantly in the cases where the probe face images' quality is very low. Finally, we experiment different face crop sizes in order to assess the impact of information included in the face images. Experimental results indicate that cropping a larger region of the face images improves the performance.

By leveraging these factors, we achieve state-of-the-art results on the SCFace~\cite{grgic2011scface} and ICB-RW~\cite{neves2016icb} benchmarks, even without using any data from these benchmarks to train or fine-tune the employed deep CNN models. To compare our results on the SCFace~\cite{grgic2011scface} benchmark with~\cite{lu2018deep}, we conduct 10 Repeated Random Sub-Sampling Validation (RRSSV) experiment on 80 subjects out of 130 subjects and report the mean and standard deviation of Rank-1 Identification Rate (IR). We achieve $78.5\% \pm 1.67$, $98.38\% \pm 0.48$, and $99.75\% \pm 0.16$ Rank-1 IR for distance 1, 2, and 3 (d1, d2, and d3) of SCFace~\cite{grgic2011scface}, respectively. Our approach outperforms the state-of-the-art results presented in Deep Coupled-ResNet (DCR)~\cite{lu2018deep} by the large margins of 5.2\%, 4.88\%, and 1.75\% for d1, d2, and d3, respectively. In contrast to DCR~\cite{lu2018deep}, we do not exploit target dataset for fine-tuning. Furthermore, we evaluate the proposed factors on ICB-RW benchmark~\cite{neves2016icb} and outperform the results reported in Ghaleb et al.~\cite{ghaleb2018deep}, the best performing system in the ICB-RW 2016 challenge~\cite{neves2016icb}, by a significant margin of 12.52\% for Rank-1 IR.

The remainder of this paper is organized as follows. In section~\ref{relatedwork}, we provide an overview of related work. In section \ref{methods}, face detection, feature extraction, and face identification steps are explained. Experimental results are presented and discussed in section \ref{results}. Finally, in section \ref{conclusions} conclusions of this work are summarized.

\section{Related Work}\label{relatedwork} 
The related works for face recognition can be grouped into Low Resolution Face Recognition (LRFR) and High Resolution Face Recognition (HRFR). The reported results on the HRFR benchmarks are nearly perfect. In FaceNet~\cite{schroff2015facenet}, a Deep Convolutional Neural Network (DCNN) architecture with Inception~\cite{szegedy2015going} modules is trained on a very large-scale database of 260 M images. After that, the features are L2 normalized and triplet loss is proposed to learn deep face representations. The proposed method achieved 99.63\% face IR on the LFW benchmark~\cite{huang2008labeled} and 95.12\% face IR on the YTF benchmark~\cite{wolf2011face}. Sun, et al. \cite{Sun2015DeepID3FR} included Inception modules \cite{szegedy2015going} into two VGG architectures~\cite{simonyan15}, and concatenated extracted features from 25 different crops of each face per network. Afterwards, a joint Bayesian model is learned for face recognition. The proposed method achieved 99.54\% verification accuracy on the LFW~\cite{huang2008labeled}. In SphereFace~\cite{liu2017sphereface} the Angular-Softmax loss is introduced and adopted ResNet architecture~\cite{he2016deep} to learn face embeddings in training phase. They applied nearest neighbor classifier with cosine similarity for face identification. The applied method achieved 99.42\% verification accuracy on the LFW~\cite{huang2008labeled} and 95.0\% on YTF~\cite{wolf2011face} datasets, respectively. ArcFace~\cite{deng2018arcface} leveraged ResNet~\cite{he2016deep} architecture and train the face identification model with additive angular margin loss. Their reported best verification accuracy are 99.83\% on the LFW~\cite{huang2008labeled} and 98.02\% on the YTF~\cite{wolf2011face}.\par

In contrast to HRFR, the performance of deep CNN models degrade significantly in LRFR. Lee et al.~\cite{lee2012local}, extracted local color vector binary patterns and nearest neighbor classifier with euclidean distance metric are carried out for face identification. Average Rank-1 Identification Rate (IR) of  67.68\% is reported for distance 1 and 2 (4.20m, 2.60m, respectively) of SCFace~\cite{grgic2011scface}. De Marsico et al.~\cite{de2013robust} applied pose and illumination normalization on faces and localized spatial correlation index for face matching. They reported 89\% Rank-1 IR for distance 3 (1.0m) of SCFace~\cite{grgic2011scface}. A Patch Based Cascaded Local Walsh Transform (PCLWT) followed by whitened principal component analysis is employed in \cite{uzun2018face} for feature extraction. They reported 64.76\%, 80.8\%, and 74.92\% Rank-1 IR for d1, d2, and d3 respectively. Yang et al.~\cite{yang2018discriminative} proposed the Local-Consistency-Preserved Discriminative Multidimensional Scaling (LDMDS) approach to learn compact intra-class features and maximize inter-class distance. They selected 50 subjects, out of 130 subjects available in SCFace~\cite{grgic2011scface}, for training and calculated Rank-1 IR for the remaining 80 subjects. They reported 62.7\%, 70.7\%, and 65.5\% Rank-1 IR for distance 1, 2, and 3, respectively. Following~\cite{yang2018discriminative}, in Deep Coupled-ResNet (DCR)~\cite{lu2018deep} a two-step multi-scale training strategy is performed to train a trunk and two branches (HR and LR branches). They trained the trunk network with three different image resolutions ($112 \times 96$, $40 \times 40$, and $6 \times 6$) pre-processed from CASIA Webface database~\cite{Yi2014LearningFR}. In the second step, they fixed the weights of the trunk network and trained the HR and LR branches. For that, they trained HR branch with $112 \times 96$ pixel resolution and LR branch with $112 \times 96$, $30 \times 30$, and $20 \times 20$ pixel resolutions based on the image resolutions of the distances 1, 2, and 3 respectively. After that, they fine-tuned the HR and LR branches with 50 randomly selected subjects of the SCFace dataset~\cite{grgic2011scface}. Deep face embedding of the gallery and probe faces of SCFace dataset~\cite{grgic2011scface} are extracted with HR branch and LR branches, respectively. They evaluated the proposed method on 80 remaining subjects of SCFace dataset~\cite{grgic2011scface} and achieved 73.3\%, 93.5\%, and 98.00\% Rank-1 IR for distance 1 (4.20m), distance 2 (2.60m), and distance 3 (1.0m), respectively. As it can be noticed from these results, performance of the proposed methods deteriorate significantly when the resolution of the probe faces decreases. In GenLR-Net~\cite{mudunuri2018genlr} authors employed VGGFace~\cite{parkhi2015deep} pre-trained model to construct two branches network to overcome performance degradation in LR face recognition. Their proposed method significantly improved the results of HR-LR verification task on modified fold 1 of LFW benchmark~\cite{huang2008labeled} from 69.16\% using original VGGFace~\cite{parkhi2015deep} model to 90.00\%. There are also deep learning based super-resolution methods to deal with low resolution faces, however, these methods are not optimized for LRFR~\cite{yu2018super} and yield modest performance improvement~\cite{wang2016studying}.

\section{Methodology}\label{methods}

In the following sections, we present the building blocks of the system, which are employed face detector \cite{zhang2016joint}, utilized training databases \cite{guo2016ms, cao2018vggface2} and the deep CNN models \cite{cao2018vggface2, deng2018arcface}, proposed strategy to match the resolution of the gallery and probe images, the crop ratios to adjust the amount of information included in the face images, and finally the similarity measurement and the evaluation metric.

\subsection{Face Detection}

The bounding boxes of the faces in the gallery and probe sets are detected using the Multi-Task Cascaded Convolutional Neural Networks (MTCNN)~\cite{zhang2016joint} model. The faces are cropped and resized to $224 \times 224$ or $112 \times 112$ pixel resolutions depending on the input size of the deep learning models. 

\subsection{Feature Extraction}

We employ four state-of-the-art deep CNNs, namely ResNet-50~\cite{he2016deep}, SENet-50~\cite{hu2018squeeze}, LResNet50E-IR~\cite{deng2018arcface}, and LResNet100E-IR~\cite{deng2018arcface}. The deep models are trained or fine-tuned on VGGFace2~\cite{cao2018vggface2} and MS-Celeb-1M databases~\cite{guo2016ms} to learn the face embedding of the gallery and probe face images in the SCFace~\cite{grgic2011scface} and ICB-RW~\cite{neves2016icb} benchmarks. Please note that we do not take advantage of these benchmarks for fine-tuning.

\subsubsection{Deep face models}
The deep face models that are utilized in this study are listed in Table \ref{tb:models} and named as model \textit{a, b, c, ..., h}. Off-the-shelf models described in VGGFace2~\cite{cao2018vggface2} and ArcFace~\cite{deng2018arcface} are used for models \textit{a, b, c, d} and \textit{e, g}, respectively. Furthermore, models \textit{e} and \textit{g} are fine-tuned on the VGGFace2 database~\cite{cao2018vggface2} to learn models \textit{f} and \textit{h}, respectively. 

\begin{table*}[!hbt]
\begin{center}
\begin{tabular}{|c|c|c|c|c|c|}
\hline
Models    & CNNs  & Trained on  & Fine-tuned on  & Input size  & Embedding size \\ \hline
\textit{a}  & ResNet-50~\cite{he2016deep} & VGGFace2~\cite{cao2018vggface2} & n/a & $224 \times 224$ & 2048  \\ \hline
\textit{b}  & ResNet-50~\cite{he2016deep} & MS-Celeb-1M~\cite{guo2016ms} & VGGFace2~\cite{cao2018vggface2} & $224 \times 224$ & 2048 \\ \hline
\textit{c}  & SENet-50~\cite{hu2018squeeze} & VGGFace2~\cite{cao2018vggface2} & n/a & $224 \times 224$ & 2048  \\ \hline
\textit{d}  & SENet-50~\cite{hu2018squeeze} & MS-Celeb-1M~\cite{guo2016ms} & VGGFace2~\cite{cao2018vggface2} & $224 \times 224$ & 2048 \\ \hline
\textit{e}  & LResNet50E-IR~\cite{deng2018arcface}  & MS-Celeb-1M~\cite{guo2016ms} & n/a & $112 \times 112$ & 512 \\ \hline
\textit{f}  & LResNet50E-IR~\cite{deng2018arcface} & MS-Celeb-1M~\cite{guo2016ms} & VGGFace2~\cite{cao2018vggface2} & $112 \times 112$ & 512 \\\hline
\textit{g}  & LResNet100E-IR~\cite{deng2018arcface} & MS-Celeb-1M~\cite{guo2016ms} & n/a & $112 \times 112$ & 512 \\ \hline
\textit{h}  & LResNet100E-IR~\cite{deng2018arcface} & MS-Celeb-1M~\cite{guo2016ms} & VGGFace2~\cite{cao2018vggface2} & $112 \times 112$ & 512 \\ \hline
\end{tabular}
\end{center}
\caption{The eight combinations resulting from the different deep CNN architectures and training databases that are used for feature extraction in this study.}
\label{tb:models}
\end{table*}

\subsubsection{Fine-tuning}

The detected face images of the VGGFace2 database~\cite{cao2018vggface2} are aligned with respect to the positions of the center of the eyes, tip of the nose, and the corners of the mouth. The aligned faces are then resized to $112 \times 112$ pixel resolution and finally pixel intensity values are normalized by subtracting 127.5 and dividing by 128. These pre-processed face images 
are then provided for fine-tuning.

\textit{Model f}: model \textit{e} is fine-tuned on the VGGFace2 database~\cite{cao2018vggface2} using additive angular margin loss~\cite{deng2018arcface} with $m=0.5$ and $s=64.0$. Stochastic gradient descent with momentum 0.9 and learning rate of 0.01 are used to fine-tune the network with the batch size of 64. The learning rate is divided by 10 at 20K, 28K iterations and the training
process is stopped at 32K iterations as in ArcFace~\cite{deng2018arcface}. The obtained verification accuracy of the validation set, LFW dataset~\cite{huang2008labeled}, is 99.6\%.

\textit{Model h}: model \textit{g} is fine-tuned on the VGGFace2 database~\cite{cao2018vggface2} with the same setting as in the model \textit{f}, however, the learning rate is set to 0.001. The achieved verification accuracy on the LFW dataset~\cite{huang2008labeled} is 99.7\%.

\subsection{Amount of information}\label{information}
\begin{figure}[!ht]
 \begin{center}
 \includegraphics[width=\linewidth,keepaspectratio=true]{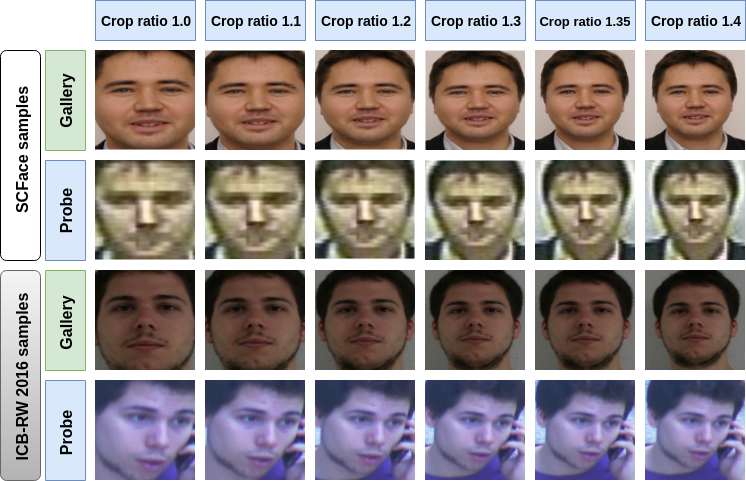}
 \end{center}
 \caption{Gallery and probe faces of a subject from SCFace and ICB-RW benchmarks cropped with six different crop ratios.}
 \label{fig:boxes}
\end{figure}

To adjust the amount of information to be included in the face images, we extend the face bonding boxes. In a previous work~\cite{mehdipour2016comprehensive}, it has been shown that this has a significant effect on the performance. In our study, we also expect this adjustment to contribute positively to the performance of LRFR due to two main reasons. The first reason is that due to low resolution, the face images contain limited information, extending face bounding boxes would allow to include more information, for example about the shape of the face, etc. The second one is related to the upsampling factor. Since input size of the face images to the deep learning models are relatively high, in our case $224 \times 224$ or $112 \times 112$ pixels, this requires upsampling of the low resolution face images with a large scaling factor. A larger crop of the face region would decrease the scaling factor, thus, less degradation would occur due to upsampling. In this work, we control the amount of information to be included in the face images with six different crop ratios (1.0, 1.1, 1.2, 1.3, 1.35, 1.40) as shown in Figure \ref{fig:boxes}.

\begin{figure}[!ht]
 \begin{center}
 \includegraphics[width=\linewidth,keepaspectratio=true]{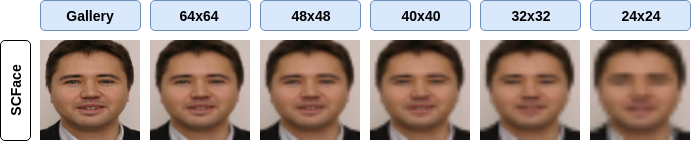}
 \end{center}
 \caption{Gallery faces of a subject from SCFace benchmark cropped with 1.3 extension factor and matched resolution of them with five different pixel resolutions ($24 \times 24$, $32 \times 32$, $40 \times 40$, $48 \times 48$, $64 \times 64$) are shown here.}
 \label{fig:downscaled}
\end{figure}

\subsection{Matching the resolution}\label{resolution}
An important challenge in LRFR is that features extracted from very low resolution faces in the probe set and high resolution images in the gallery set can potentially have higher intra-class distance than inter-class distance. We hypothesize that if we could make the appearance of the gallery face images similar to the probe face images, intuitively, we would minimize the intra-class distance. Therefore, to imitate low resolution we downsampled the gallery images. That is the gallery face images are downsampled and this way their resolution is matched with the resolution of probe face images. For this purpose we picked five different resolutions ($24 \times 24$, $32 \times 32$, $40 \times 40$, $48 \times 48$, $64 \times 64$) and select $32 \times 32$, $48 \times 48$, and $64 \times 64$, which are closest to the resolution of d1, d2, and d3 probe face images in SCFace~\cite{grgic2011scface}, respectively. We take the original resolution of gallery face images in ICB-RW~\cite{neves2016icb} experiments, which matches the resolution of the probe face images. In Figure \ref{fig:downscaled}, the first column shows the gallery face of a subject from SCFace~\cite{grgic2011scface} cropped with $1.3$ extension ratio, whereas, the other columns show downsampled gallery face images at five different resolutions to make their image quality similar to the probe face images in the SCFace~\cite{grgic2011scface} benchmark.\par

\subsection{Face Identification}
The face embedding of the gallery and probe sets are extracted using eight deep face models described in Table~\ref{tb:models}. The identification task for probe faces are carried out by nearest neighbor classification method with the correlation distance metric (eq. \ref{eq:corr})
as similarity measurement:  
\begin{equation}
\label{eq:corr}
Corr.distance(u,v) = 1 - \frac{(u-\bar{u}).(v-\bar{v})}{\Vert(u-\bar{u})\Vert_{2}\Vert(v-\bar{v})\Vert_{2}}
\end{equation}
where \textit{u, v} are the face feature vectors and $\bar{u}, \bar{v}$ are mean of the face feature vectors. Rank-1 IR is reported as the evaluation metric.

\section{Experimental Results}\label{results}
We conduct our experiments in three steps on the SCFace and ICB-RW benchmarks. Firstly, we crop faces with bounding boxes detected by MTCNN~\cite{zhang2016joint} before feature extraction. Secondly, larger crops are used for feature extraction, and finally, the gallery faces' pixel resolution are matched with the resolution of probe face images before extracting the face embedding. In this section, we provide the experimental results for these steps.

\subsection{Datasets}
We evaluate the proposed methods on the SCFace~\cite{grgic2011scface} and ICB-RW~\cite{neves2016icb} benchmarks. 

There are 130 subjects in SCFace dataset~\cite{grgic2011scface}, one frontal image (gallery set) and 15 LR images per subject (probe set). The gallery faces are captured in controlled conditions, whereas, the probe faces are captured with five indoor surveillance cameras located at three different distances, d1, d2, and d3 (4.20, 2.60, and 1.00 meters, respectively) resulting in the probe images with varying image quality. Please note that in this study we do not fine-tuned our models with target dataset and we report the Rank-1 IR for 130 subjects of SCFace~\cite{grgic2011scface}. However, in order to be able to compare our results with previous works, we report the mean and standard deviation of Rank-1 IR of 10 RRSSV experiments for 80 subjects out of 130 in model \textit{h}* (Table \ref{tb:matching}).

ICB-RW benchmark \cite{neves2016icb} contains 90 subjects, each having one high quality gallery image and 5 probe images, recorded outdoors, containing variations in illumination, expression, pose, motion-blur, occlusion, and focus. Figure~\ref{fig:probe-samples} illustrates the aforementioned probe image quality problems in SCFace~\cite{grgic2011scface} and ICB-RW~\cite{neves2016icb} benchmarks.

\subsection{Baseline experiments}\label{original}
The faces are detected using the MTCNN~\cite{zhang2016joint} and cropped according to the face detection output. The face embeddings are extracted with eight deep CNN models, as presented in Table~\ref{tb:models}. Thereupon, face embeddings are fed into the nearest neighbor classifier with correlation distance metric as the similarity measurement. The Rank-1 IR results on the SCFace~\cite{grgic2011scface} and ICB-RW~\cite{neves2016icb} benchmarks are reported in Table~\ref{tb:mtcnn}. It can be seen from the results that the performance of the state-of-the-art deep CNN models plummet at d1, which contains very low resolution probe face images. We fine-tune models \textit{e} and \textit{g} using VGGFace2 database~\cite{cao2018vggface2} to learn models \textit{f} and \textit{h}, respectively. After that, a significant improvement in performance of models \textit{f} and \textit{h} for d1 of SCFace~\cite{grgic2011scface} (see Table~\ref{tb:mtcnn}) are observed. The improvement can be described to the fact that the VGGFace2 database~\cite{cao2018vggface2} contains approximately 20\% of the face images with pixel resolution lower than 50 pixel, which allow the model to learn better feature representation for low resolution face images.
\begin{table}[!ht]
\begin{center}
\begin{tabular}{c|ccc|c}
\hline
 &    \multicolumn{3}{c|}{SCFace}      & ICB-IRW  \\ \hline
Model & d1       & d2       & d3       & probe  \\ \hline
\textit{a}     & 40.15 & \textbf{91.38} & \textbf{98.15} & 79.11 \\ \hline
\textit{b}     & \textbf{41.85} & 89.54 & 97.69 & 77.56 \\ \hline
\textit{c}    & 33.08 & 86.92 & 96.62 & \textbf{81.33} \\ \hline
\textit{d}     & 35.69 & 86.00 & 97.23 & 79.56 \\ \hline
\textit{e}     & 13.85 & 59.54 & 86.31 & 40.44 \\ \hline
\textit{f}     & 20.46 & 71.54 & 85.38 & 48.00 \\ \hline
\textit{g}     & 25.38 & 84.00 & \textbf{98.15} & 68.22 \\ \hline
\textit{h}     & 37.54 & 87.69 & 96.00 & 69.33 \\ \hline
\end{tabular}
\\
\end{center}
\caption{The Rank-1 IR results (\%) of eight deep models are reported for d1 (4.2 m), d2 (2.6 m), and d3 (1.0 m) probe faces of SCFace and ICB-RW in which we detected the faces with MTCNN model and cropped them with 1.0 ratio.}
\label{tb:mtcnn}
\end{table}

\begin{figure*}[!ht]
 \begin{center}
 \includegraphics[width=\linewidth,keepaspectratio=true]{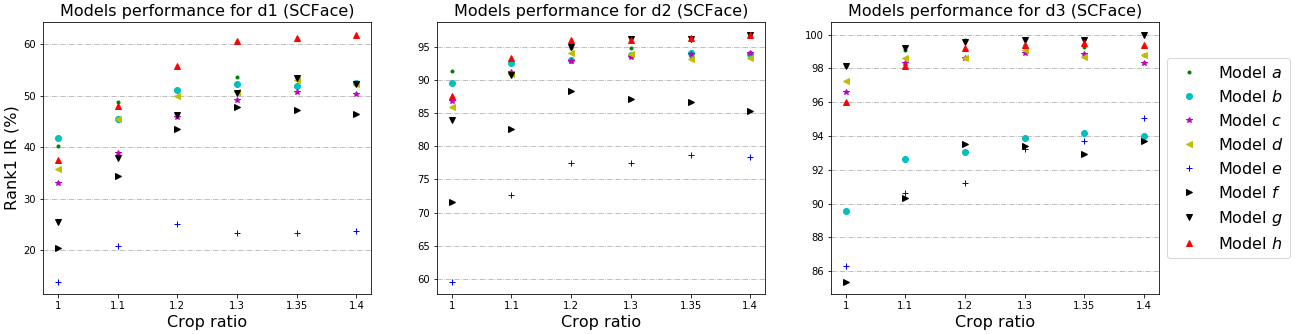}
 \end{center}
 \caption{The Rank-1 IR (\%) of deep CNN models on probe faces of SCFace benchmark for six different crop ratios.}
 \label{fig:cropratio-scface}
\end{figure*}

\begin{figure}[!ht]
 \begin{center}
 \includegraphics[width=\linewidth,keepaspectratio=true]{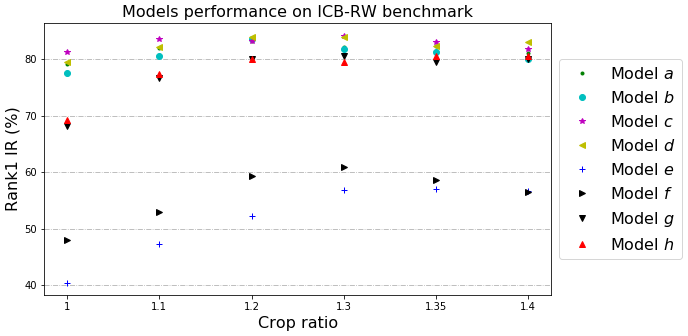}
 \end{center}
 \caption{The Rank-1 IR (\%) of deep CNN models on probe faces of ICB-RW benchmark for six different crop ratios.}
 \label{fig:cropratio-icbrw}
\end{figure}

\begin{table}[!ht]
\begin{center}
\begin{tabular}{c|ccc|c}
\hline
 &    \multicolumn{3}{c|}{SCFace}      & ICB-IRW  \\ \hline
Model & d1       & d2       & d3       & probe  \\ \hline
\textit{a}     & 53.54 & 94.92 & 99.38 & 80.67 \\ \hline
\textit{b}     & 52.15 & 93.85 & 98.00 & 81.56 \\ \hline
\textit{c}     & 49.08 & 93.54 & 98.92 & 82.00 \\ \hline
\textit{d}     & 50.77 & 94.00 & 99.08 & \textbf{82.67} \\ \hline
\textit{e}     & 23.23 & 77.54 & 93.23 & 58.22 \\ \hline
\textit{f}     & 47.69 & 87.23 & 93.38 & 60.00 \\ \hline
\textit{g}     & 50.46 & \textbf{96.31} & \textbf{99.69} & 82.00 \\ \hline
\textit{h}     & \textbf{60.62} & 96.15 & 99.38 & 78.67 \\ \hline
\end{tabular}
\\
\end{center}
\caption{The Rank-1 IR (\%) of deep CNN models using 1.30 crop ratio are reported for d1, d2, and d3 in SCFace and probe faces of ICB-RW.}
\label{tb:cropratios}
\end{table}

\begin{table}[!ht]
\begin{center}
\begin{tabular}{c|ccc|c}
\hline
 &    \multicolumn{3}{c|}{SCFace}      & ICB-RW  \\ \hline
Model & d1       & d2       & d3       & probe  \\ \hline
\textit{a}     & 56.72 & 95.23 & 99.23 & 82.22 \\ \hline
\textit{b}     & 59.38 & 96.00 & 98.00 & 82.00 \\ \hline
\textit{c}     & 54.15 & 94.77 & 98.92 & \textbf{84.22} \\ \hline
\textit{d}     & 60.62 & 94.46 & 99.23 & 84.00 \\ \hline
\textit{e}     & 33.38 & 80.62 & 95.23 & 58.67 \\ \hline
\textit{f}     & 55.38 & 89.69 & 93.85 & 60.89 \\ \hline
\textit{g}     & 67.08 & 97.23 & \textbf{100} & 81.78 \\ \hline
\textit{h}     & 75.08 & 97.69 & 99.69 & 79.78 \\ \hline
\textit{h}*     & \textbf{78.5} & \textbf{98.38} &  99.75 & n/a \\ \hline
DCR~\cite{lu2018deep}     & 73.3 & 93.5 & 98.0 & n/a \\ \hline
LDMDS~\cite{yang2018discriminative}     & 62.7 & 70.7 & 65.5 & n/a \\ \hline
PCLWT~\cite{uzun2018face}    & 64.76 & 80.8 & 74.92 & n/a \\ \hline
Ghaleb et al.~\cite{ghaleb2018deep}     & n/a & n/a & n/a & 71.7 \\ \hline
\end{tabular}

\end{center}
\caption{The results achieved with 1.3 crop ratio are reported for DCNN models. * denotes that model \textit{h}* results are mean of 10 RRSSV for 80 subjects out of 130 in SCFace~\cite{grgic2011scface}. The presented mean face identification rates for d1, d2, and d3 have 1.67, 0.48, and 0.16 standard deviation, respectively.}
\label{tb:matching}
\end{table}

\subsection{Effect of increasing the amount of information}\label{ext}
As we discussed in section \ref{information}, we control the amount of information to be included in the gallery and probe face images by using six different crop ratios. Empirical results show a compelling improvement on the performance of eight deep CNN models. We plot the Rank-1 IR of deep CNN models for each of six crop ratios as illustrated in Figure \ref{fig:cropratio-scface} for SCFace~\cite{grgic2011scface}, and Figure \ref{fig:cropratio-icbrw} for ICB-RW~\cite{neves2016icb} benchmarks. Table \ref{tb:cropratios} summarizes the Rank-1 IR results achieved by 1.30 crop ratios for the eight deep models. These results show the impact of the increased information in the significant improvement of the models' performance, especially, for the probe face images that have lower resolution. Our results also validate the results in \cite{mehdipour2016comprehensive}, which presented the performance improvement in face recognition using extended bounding boxes.

\subsection{Effect of matching the resolution}\label{down}
As it is mentioned in section \ref{resolution}, we conduct experiments on SCFace~\cite{grgic2011scface} and ICB-RW~\cite{neves2016icb} benchmarks using eight deep CNN models to test the contribution of matching the resolution at performance improvement. We observe that Rank-1 IR improves significantly for the low resolution probe faces as in SCFace~\cite{grgic2011scface}, however, there is not much improvement in the higher resolutions probe images as in ICB-RW~\cite{neves2016icb} which already have a matching resolution with the gallery face images. Table \ref{tb:matching} shows the Rank-1 IR achieved by DCNN models on SCFace~\cite{grgic2011scface} and ICB-RW benchmark~\cite{neves2016icb}. The models with $224 \times 224$ input size (\textit{a}, \textit{b}, \textit{c}, \textit{d}) achieve higher Rank-1 IR for ICB-RW benchmark~\cite{neves2016icb}, which can be described to the fact that the probe images of ICB-RW~\cite{neves2016icb} have higher resolution. The presented Rank-1 IR on SCFace benchmark~\cite{grgic2011scface} are achieved with $32 \times 32$, $48 \times 48$, and $64 \times 64$ downsampled gallery face images which are close to the average resolution of d1, d2, and d3 in SCFace benchmark~\cite{grgic2011scface}, respectively. Please note that in DCR~\cite{lu2018deep} and LDMDS~\cite{yang2018discriminative} randomly selected 50 subjects out of 130 subjects in SCFace~\cite{grgic2011scface} are used for fine-tuning and the results are reported on 80 remaining subjects. To compare our results we also report the mean and standard deviation of 10 RRSSV experiments on 80 randomly selected subjects (model \textit{h}*). As can be seen from Table \ref{tb:matching}, on d1 and d2 subsets around 5\% and on d3 subset 2\% absolute performance improvement has been achieved compared to the DCR~\cite{lu2018deep} leading to the state-of-the-art results for the SCFace dataset~\cite{grgic2011scface}. Similarly, the proposed approach enhances the state-of-the-art accuracy on the ICB-RW benchmark~\cite{neves2016icb} from 71.7\% to 84.22\%. \par

\section{Conclusion}\label{conclusions}
In this paper, we explore the factors that would contribute to improve identification accuracy of low resolution face recognition under  mismatched conditions. We observe that models \textit{f} and \textit{h} fine-tuned on the VGGFace2 dataset significantly improve Rank-1 IR for very low resolution probe face images (d1 of SCFace) compared to off-the-shelf models (models \textit{e} and \textit{g}), which are trained on MS-Celeb-1M dataset~\cite{guo2016ms}. This can be explained to the fact that VGGFace2~\cite{cao2018vggface2} has about 20\% of the face images with resolution lower than 50 pixels, which helps the model to learn robust features for low resolution faces. The experimental results show that including more information in the cropped faces and matching the resolution between gallery and probe sets enhance the Rank-1 IR significantly. Our model \textit{h} achieves state-of-the-art Rank-1 IR results on 130 subjects of SCFace benchmark~\cite{grgic2011scface} which are 75.08\%, 97.69\%, and 99.69\% Rank-1 IR for d1, d2, and d3 respectively. We also significantly improve the Rank-1 IR on ICB-RW benchmark with model \textit{c} that achieves 84.22\% Rank-1 IR outperforming the validation results reported in Ghaleb et al~\cite{ghaleb2018deep} by 12.52 margin.

{\small
\bibliographystyle{ieee}

}
\end{document}